\documentclass[10pt,twocolumn,letterpaper]{article}
\usepackage{iccv}
\usepackage{times}
\usepackage{amsmath}
\usepackage{amssymb}
\usepackage{bm}
\usepackage{booktabs}
\usepackage{graphicx}
\usepackage{graphicx}
\usepackage{microtype}
\usepackage{nicefrac}
\usepackage{url}
\usepackage[pagebackref=true,breaklinks=true,letterpaper=true,colorlinks,bookmarks=false]{hyperref}
\usepackage{cleveref}

\graphicspath{{./figures/}}

\renewcommand{\b}[1]{{\textbf{#1}}}
\renewcommand{\u}[1]{{\underline{#1}}}
\renewcommand{\d}{${}^\dagger$}
\renewcommand{\t}[1]{{\text{#1}}}
\newcommand{\pkp}{p_{\t{keep\_patch}}}
\newcommand{\pki}{p_{\t{keep\_image}}}

\newcommand\blfootnote[1]{%
	\begingroup
	\renewcommand\thefootnote{}\footnote{#1}%
	\addtocounter{footnote}{-1}%
	\endgroup
}

\iccvfinalcopy
\def\iccvPaperID{7}

\ificcvfinal\fi
\begin{document}

\title{Occlusions for Effective Data Augmentation in Image Classification}

\author{
  Ruth C. Fong\thanks{Work done as a contractor at FAIR.} \\
  University of Oxford \\
  \and
  Andrea Vedaldi \\
  Facebook AI Research
}

\maketitle

\begin{abstract}
Deep networks for visual recognition are known to leverage ``easy to recognise'' portions of objects such as faces and distinctive texture patterns.
The lack of a holistic understanding of objects may increase fragility and overfitting.
In recent years, several papers have proposed to address this issue by means of occlusions as a form of data augmentation.
However, successes have been limited to tasks such as weak localization and model interpretation, but no benefit was demonstrated on image classification on large-scale datasets.
In this paper, we show that, by using a simple technique based on batch augmentation,  occlusions as data augmentation can result in better performance on ImageNet for high-capacity models (e.g., ResNet50).
We also show that varying amounts of occlusions used during training can be used to study the robustness of different neural network architectures.
\end{abstract}

\section{Introduction}\label{s:introduction}

Robustness to occlusions is an important property of image recognition systems.
That is, a robust image classifier should be able to solve the problem even if only a portion of the object of interest is visible in an image.
However, the image classification datasets commonly used to train high-performance models such as deep neural networks are strongly affected by the so called ``photographer bias''.
Among other things, this bias means that the main subject of these pictures tends to be centred and clearly visible.
As a consequence, learning a model on such data may results in ``lazy'' networks that focus too much on easily recognizably details (such as the face of a cat) and cannot understanding other, more subtle cues (such as the cat's body) that may be important in harder scenarios.

A few authors have proposed to address this issue by augmenting the training data via simulated occlusions.
While details change depending on the specific method, the general idea is that, if part of the image is not visible at training time, then the network should be stimulated to learn to recognize all available evidence, thus avoiding to over-rely on the most obvious evidence.
However, the success of these techniques has been mixed.
\cite{wei2017object,singh2017hide} showed improvements on the ability of the network to localize objects but not on the original task of object recognition.
\cite{devries2017improved} demonstrated better performance in classification performance in simpler datasets such as CIFAR10~\cite{krizhevsky2014cifar} but not in larger, more complex ones such as ImageNet~\cite{russakovsky2015imagenet} (as confirmed in our experiments and in~\cite{ghiasi2018dropblock}).

A hypothesis for this behaviour is that training using occlusion augmentation improves the robustness of the model to occlusions, but that this does not correspond to a test-time performance improvement because the test set does not, in fact, contain occlusions.

In this paper, we show that this is not the case.
The issue can be solved, and a performance improvement observed consistently, provided that the augmentation is incorporated properly in the training procedure.
We make three main contributions:
(1) We demonstrate that augmenting image batches with several versions of the same image, in the spirit of \emph{batch augmentation}~\cite{hoffer2019augment}, allows occlusion augmentation to consistently outperform the baselines on for CNN architectures that are sufficiently powerful (e.g., ResNet50).
(2) We present a detailed analysis of why occlusion augmentation has not yielded improvements in the past.
(3) We conduct a thorough investigation on how to optimally tune occlusion augmentation, showing differences as a function of the model architecture.
For example, we demonstrate that more powerful models (e.g., ResNet50~\cite{he2016deep}) can handle, and benefit from, significantly more substantial occlusions during training than weaker ones (e.g., AlexNet~\cite{krizhevsky2012imagenet}).

\section{Related work}\label{s:related}

Occlusions have been successfully used for model interpretability and weak localization.
A few attribution methods have used fixed~\cite{zeiler2014visualizing}, stochastic~\cite{petsiuk2018rise}, and optimized~\cite{fong17interpretable} occlusions to diagnose ``where'' a network is ``looking'' in the input for evidence for its prediction.
A few works have demonstrated that applying random~\cite{singh2017hide} or optimized~\cite{wei2017object} occlusions to the input or intermediate activations~\cite{wang2017fast} can improve weakly supervised localization (but not necessarily image classification) by forcing a classification network to be robust to occlusions and thus rely other parts of an object besides its most discriminative parts.

Cutout~\cite{devries2017improved} and Hide-and-Seek~\cite{singh2017hide} both introduce stochastic input-level occlusions:
Cutout ``drops'' (i.e., zeros out) randomly positioned squares (\Cref{f:cutout_diagram}), while Hide-and-Seek divides an image into a square grid and ``drops'' grid patches independently (\Cref{f:hide_and_seek_diagram}).
Hide-and-Seek~\cite{singh2017hide} highlights its improvements of weak localization at the expense of classification performance on ImageNet~\cite{russakovsky2015imagenet}.
Although Cutout~\cite{devries2017improved} improves performance on CIFAR10 and CIFAR100~\cite{krizhevsky2014cifar},~\cite{ghiasi2018dropblock} reported that it did not improve classification performance on ImageNet.

Other regularization methods related to occlusions are techniques inspired by Dropout~\cite{srivastava2014dropout}, which ``drop'' parts of intermediate activation tensors, such as DropPath~\cite{larsson2016fractalnet}, Scheduled DropPath~\cite{zoph2018learning}, Spatial Dropout~\cite{tompson2015efficient} and DropBlock~\cite{ghiasi2018dropblock}.
Whereas Dropout~\cite{srivastava2014dropout} drops a single voxel from a 3D activation tensor of a given input, DropPath~\cite{larsson2016fractalnet} drops a whole branch of a network while Spatial Dropout~\cite{tompson2015efficient} drops a whole slice in a 3D activation tensor associated to a filter.
DropBlock~\cite{ghiasi2018dropblock} can be viewed as an extension of Cutout~\cite{devries2017improved} applied to intermediate activations.
In this method, contiguous blocks in each activation slice associated to a filter are dropped.
These techniques, particularly DropBlock~\cite{ghiasi2018dropblock}, yield modest but consistent improvements in ImageNet~\cite{russakovsky2015imagenet} classification performance; however, they all require architectural change and, in the case of Scheduled DropPath~\cite{zoph2018learning} and DropBlock~\cite{ghiasi2018dropblock}, requires using a training schedule specific for its modules.
Label smoothing~\cite{szegedy2016rethinking} is another related regularization technique, in which noise is added to the training labels.

Recently, batch augmentation~\cite{hoffer2019augment} was introduced as a way to augment existing data augmentation techniques by including multiple copies of the same image (i.e.~copying an original batch $M$ times) and applying data augmentation to each of the copies.
When coupled with Cutout~\cite{devries2017improved}, batch augmentation significantly improved performance on small datasets like CIFAR10 and CIFAR100~\cite{krizhevsky2014cifar}.

Similar to~\cite{wei2017object}, which uses CAM (class activation maps)~\cite{zhou2016learning}, we explore using the heatmaps produced by attribution methods to occlude images during training.
We focus on the gradient-based saliency method introduced in~\cite{rebuffi2019}, which is closely related to Grad-CAM~\cite{selvaraju17gradcam} and the linear approximation~\cite{olah2018building} at a specific layer.
~\cite{rebuffi2019} shows that their method, when used to aggressively occlude images during training outperforms other baselines, including Grad-CAM~\cite{selvaraju17gradcam}.
While~\cite{rebuffi2019} focused on dataset compression and willingly sacrificed on task performance, we are interested in using occlusions to improve task performance.

Our work most directly builds off of Cutout~\cite{devries2017improved}, Hide-and-Seek~\cite{singh2017hide}, the gradient-based saliency method in~\cite{rebuffi2019}, and Batch Augmentation~\cite{hoffer2019augment}.

\section{Method}
\label{s:method}

We introduce a simple paradigm for using occlusions effectively as data augmentation.
For every image $\bm{x} \in \mathbb{R}^{3\times H \times W}$, we generate a pattern of occlusion $\bm{m} \in \mathbb{R}^{1\times H \times W}$ using one of the methods described below.
Then, for a given batch of images $\bm{X} \in \mathbb{R}^{B \times 3\times H \times W}$, we copy the batch.
We apply the set of occlusions, $\bm{M} \in \mathbb{R}^{B \times 1\times H \times W}$, to one copy of the batch, leaving the other batch unoccluded, and \emph{train jointly} with one combined batch: $(\bm{X}, \bm{X} \odot \bm{M})$.
We occlude a pixel by replacing it with the mean average colour (i.e.~setting it to zero after mean normalization).
Our joint training is inspired by batch augmentation~\cite{hoffer2019augment}.

\begin{figure}
\centering
\includegraphics[width=0.75\linewidth]{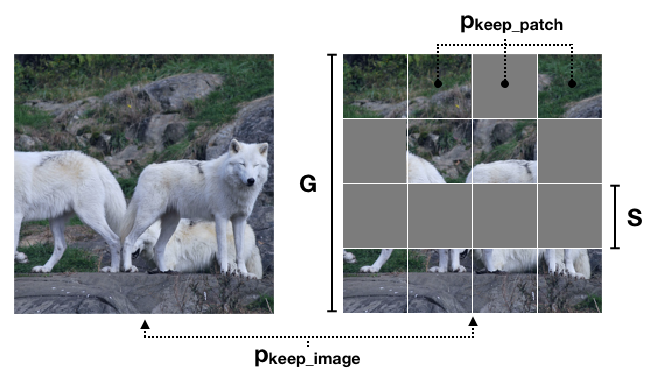}
\caption{\b{Stochastic Hide-and-Seek~\cite{singh2017hide} occlusions.}
With probability $\pki$, the image is fully preserved (left).
With $1-\pki$, each cell in a disjoint $G \times G$ grid (with side length $S$) is randomly occluded with probability $1-\pkp$ (right).
For joint training, an image duplicated into two copies, where one is always occluded ($\pki=0$) and the other never occluded ($\pki=1$).
White grid lines are used for illustration.
}\label{f:hide_and_seek_diagram}
\end{figure}

\begin{figure}
\centering
\includegraphics[width=\linewidth]{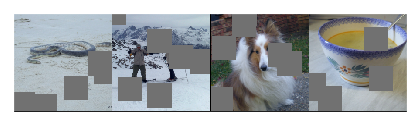}
\caption{\b{Stochastic Cutout~\cite{devries2017improved} occlusions.}
For a given image, the center points of $N$ square occlusions of side length $S$ are independently and randomly placed ($N=6, S=56$ in these examples).
}\label{f:cutout_diagram}
\end{figure}

\begin{figure}
\centering
\includegraphics[width=\linewidth]{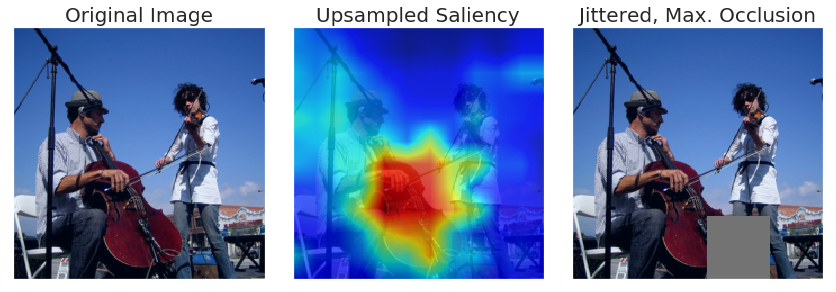}
\caption{\b{Saliency-based~\cite{rebuffi2019} occlusions.}
A saliency map at a given layer is generated according to~\Cref{e:saliency} and is then bilinearly upsampled to image size (middle).
Next, the $S \times S$ maximal patch is extracted and jittered by $\tau$ (right).
Here, a ``cello'' saliency map is generated at VGG16-BN's conv5, with $S=56$ and $\tau \in [-16,16]$.
}\label{f:saliency_diagram}
\end{figure}

\paragraph{Stochastic occlusions.} We first consider two existing ways of generating occlusions stochastically: Hide-and-Seek~\cite{singh2017hide} and Cutout~\cite{devries2017improved}.
\emph{Hide-and-Seek (H\&S)} divides an image into a $G \times G$ grid and drops patches in the grid independently with probability $1-\pkp$, where $\pkp \in [0,1]$ denotes the probability of preserving the original patch (\Cref{f:hide_and_seek_diagram}).
\emph{Cutout (CO)} drops $N$ square patches\footnote{The original Cutout paper~\cite{devries2017improved} only considers $N=1$ patch.}  of side length $S$; the center of these patches are placed uniformly at random on the whole image, thereby allowing for some patches to ``overflow'' off the image, as done in~\cite{devries2017improved} (\Cref{f:cutout_diagram}).

To analyse whether jointly training with an image occluded and unoccluded in the same batch is necessary, we introduce another hyperparameter, $\pki \in [0,1]$.
When using Hide-and-Seek occlusions without joint training (i.e., every image is in the batch exactly once), we show the full image with probability $\pki$.
Otherwise, we show an image occluded with Hide-and-Seek-style dropout, in which a patch is preserved with probability $\pkp$.
When $\pki = 0$, every image is potentially occluded; when $\pki = 1$, all images are unoccluded (i.e., standard training).

When comparing these two types of stochastic methods, Hide-and-Seek allows us to more easily and precisely define the amount of occlusion being applied on average.
This is because Cutout occlusions are allowed to flow over image boundaries and can overlap with one another in the case of $N > 1$ patches being cut out.
Pairing Hide-and-Seek with standard data augmentation (i.e., random cropping and resizing) simulates dynamic occlusions while its disjoint grid makes it easy to reason about the occlusions being applied.
Nevertheless, Cutout is more comparable to the next type of occlusions we consider: saliency-based occlusions.

\paragraph{Saliency-based occlusions.} We also consider generating occlusions based on saliency.
Given a saliency heatmap, we extract an occlusion that is most salient compared to other potential patches.
In this way, we use saliency heatmaps to guide occlusion locations as opposed to randomly sampling their locations.
This allows us to fairly compare against Cutout~\cite{devries2017improved} as we consider occlusions of the same size.

In our experiments, we use~\cite{rebuffi2019}'s gradient-based saliency method, which we summarize here (\Cref{f:saliency_diagram}; see~\cite{rebuffi2019} for more details).
For a given layer $l$, a saliency heatmap $\bm{s} \in \mathbb{R}^{H_{l} \times W_{l}}$ can be generated by computing the Frobenius norm of the product of layer $l$'s activation and gradient vectors, $\bm{x}'_{i,j}, \bm{g}_{i,j} \in \mathbb{R}^{K_l}$, at every spatial location $(i,j)$:
\begin{equation}
    s_{i,j} =  \left\lVert \bm{g}_{i,j} {\bm{x}'}_{i,j}^\top
    \right\rVert_{\text{F}} = \left\lVert \bm{g}_{i,j}\right\rVert \cdot \left\lVert\bm{x}'_{i,j} \right\rVert
\label{e:saliency}
\end{equation}
Intuitively,~\cite{rebuffi2019}'s saliency method precisely characterizes the contribution of every spatial location to the gradient of a hypothetical, subsequent $1\times 1$ convolution weights tensor initialized with identity.
We chose to use~\cite{rebuffi2019} because it generates high-quality, dense saliency maps at any network depth.
In contrast, Grad-CAM~\cite{selvaraju17gradcam} only works at the last conv layer.

For every image, we compute a saliency map with respect to the ground truth label and upsample the saliency map to the original image resolution $\mathbb{R}^{H \times W}$.
We then find the square patch with side length $S$ of the upsampled saliency map\footnote{In practice, we do this by convolving the saliency map with a $S \times S$ convolutional filter with stride $T$ and filled with 1s.}. Finally, we add a small amount of jitter $\tau$ to the extracted patches. Unlike Cutout, we do not allow our patch to overflow the image boundaries (i.e., it will always be fully contained in the image).

\section{Experiments}\label{s:experiments}

\subsection{Implementation details}
All models were trained for 100 epochs with the learning rate decayed by $0.1$ every 30 epochs (i.e., at 30, 60, and 90 epochs).
The initial learning rate for ResNet50~\cite{he2016deep} and VGG16-BN was $0.1$; for AlexNet~\cite{krizhevsky2012imagenet} and VGG16~\cite{Simonyan15} it was $0.01$\footnote{This was chosen for the non-batch normalization models based on grid search over the following learning rates: $0.1$, $0.05$, $0.01$, $0.005$, $0.001$.}.
All models used an original batch size of 256; jointly trained models used an actual batch size of 512, in which the original batch is duplicated and one copy is occluded.
The actual batch was split across 8 GPUs.
The original batch was preprocessed using standard data augmentation\footnote{We used default PyTorch ImageNet preprocessing: \footnotesize{\url{https://github.com/pytorch/examples/tree/master/imagenet}}}: random cropping to $224 \times 224$, horizontal flipping, data normalization to $\mu = 0, \sigma=1$.

When jointly training, the standard data augmentation (i.e., random cropping, etc.) occurs before the batch is duplicated, so the images are identical except for the regions that are occluded, i.e., $\bm{M}$. This differs from batch augmentation, in which images are preprocessed \emph{independently} rather than \emph{identically}.

\paragraph{Baselines.} For non-joint training baselines, we trained networks in the usual fashion without occlusions.
We introduced another set of baselines to account for the possible effect of doubling training time via joint training.
The joint training baseline refers to networks that have been trained without occlusions but with duplicated batches, that is, every image appears exactly twice in the batch.

\begin{figure*}[t]
	\centering
	\includegraphics[width=0.49\linewidth]{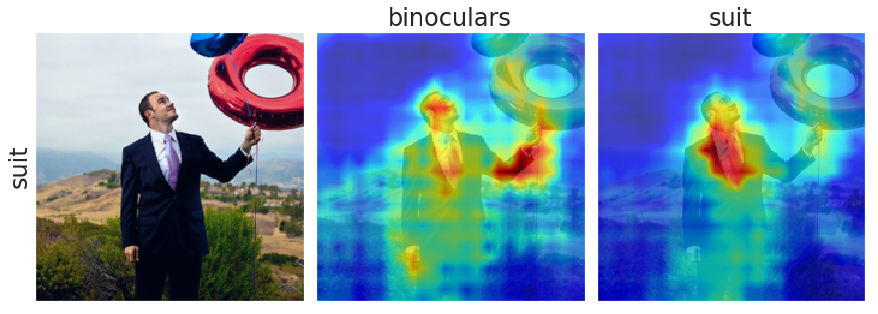}
	\includegraphics[width=0.49\linewidth]{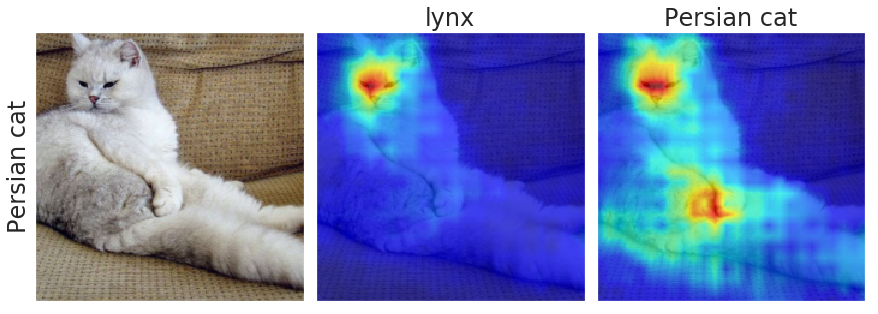}
	\includegraphics[width=0.49\linewidth]{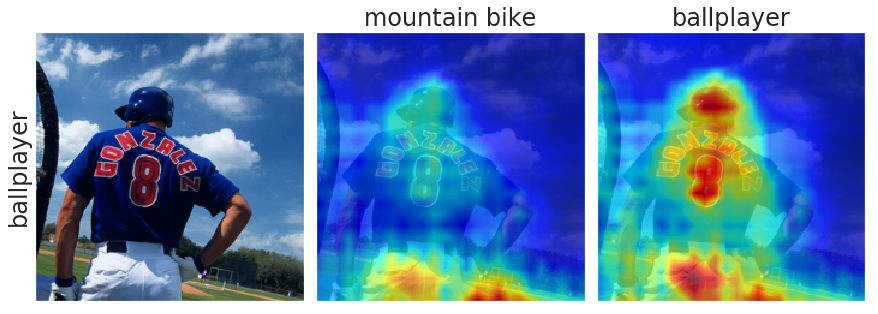}
	\includegraphics[width=0.49\linewidth]{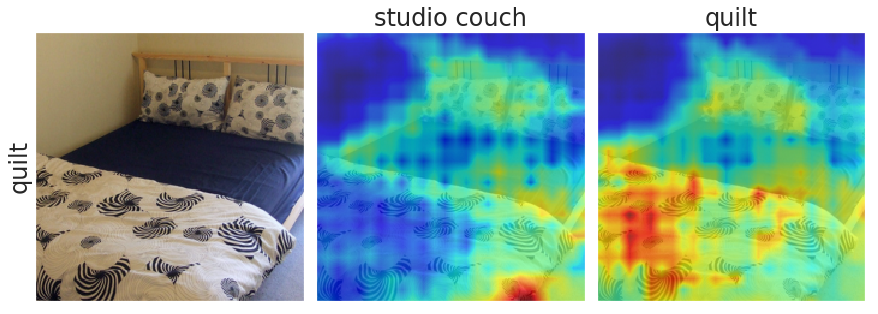}
	\includegraphics[width=0.49\linewidth]{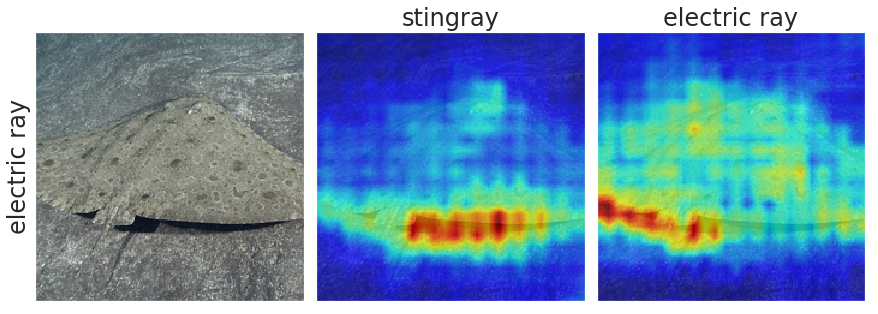}
	\includegraphics[width=0.49\linewidth]{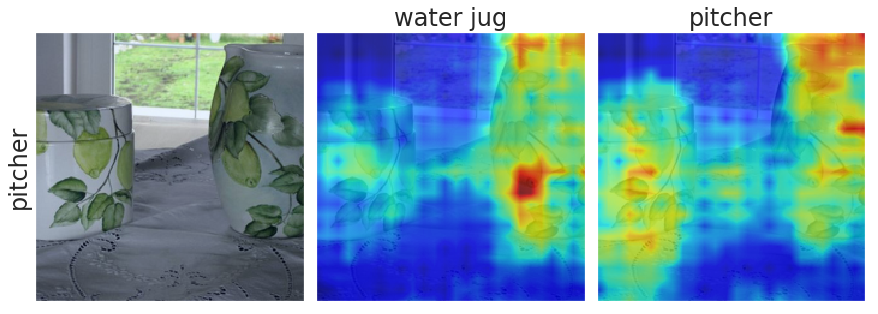}
	\includegraphics[width=0.49\linewidth]{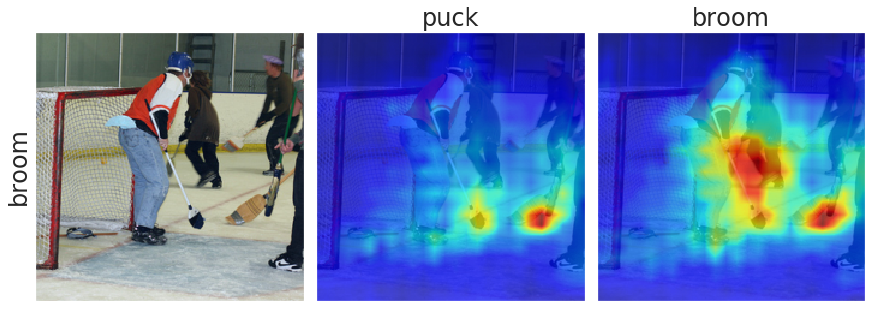}
	\includegraphics[width=0.49\linewidth]{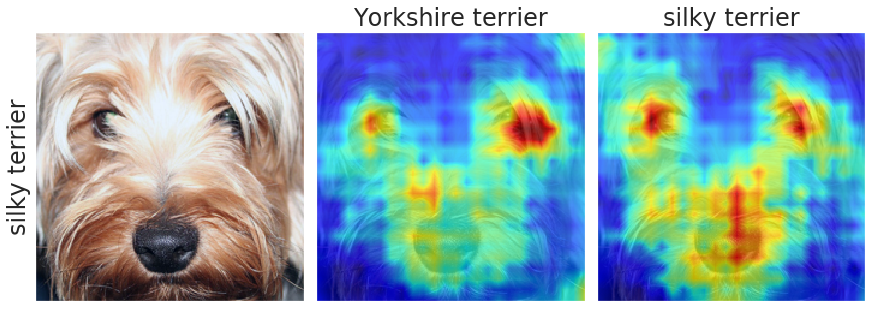}
	\caption{\b{Saliency visualizations~\cite{rebuffi2019} comparing standard vs.~occlusion-augmented ResNet50s (layer3.0.conv1).}
		Left: original image; Middle: visualization for a ResNet50 baseline without joint training ($76.43\%$ [top-1] $93.17\%$ [top-5]); Right: visualization for a ResNet50 trained jointly with Hide-and-Seek ($\pkp=0.6$; $76.43\%$ [top-1] and $93.17\%$ [top-5]).
		Top predicted classes by the network are above; ground truth labels are on the left.}\label{f:examples}
\end{figure*}

\subsection{Stochastic occlusions}

\paragraph{Experiment set-up.} We first trained networks using Hide-and-Seek~\cite{singh2017hide} occlusions.
For these experiments, we divided images into $4 \times 4$ grids ($G=4$) and preserved patches with $p \in \{0.5, 0.6, \ldots, 0.9\}$.
With a $224 \times 224$ cropped image size and grid size of $G=4$, the size of the Hide-and-Seek patches were $56 \times 56$ ($S = 56$).

Based on our Hide-and-Seek results, we then train select networks (ResNet50 and AlexNet) using Cutout-style occlusions.
Here, we occluded an image with either $N \in \{1, 2, 4, 6, 8\}$ square patches with side lengths $S \in \{56, 84, 112\}$.

For both kinds of occlusions, we trained networks jointly, that is, every batch was doubled and one copy was preserved as is (i.e., with full images) while occlusions were applied to the other copy.
At evaluation time, no occlusions are applied.

\begin{table*}[t]
	\centering
	\begin{tabular}{lcccccccc}
		\toprule
		\multicolumn{1}{c}{} & \multicolumn{2}{c}{ResNet50} & \multicolumn{2}{c}{VGG16-BN} & \multicolumn{2}{c}{VGG16} & \multicolumn{2}{c}{AlexNet} \\
		&top-1&top-5&top-1&top-5&top-1&top-5&top-1&top-5        \\ \midrule
		baseline (w/o joint)&76.40&93.10&74.11&91.81&71.75&90.45&56.39&79.19 \\
		baseline (w/ joint)&76.40&93.03&74.95&92.31&72.37&\b{90.91}&\u{57.34}&\b{79.72} \\ \midrule
		$\pkp=0.9$&76.58&93.19&74.66*&92.29*&72.20&90.73&\b{57.36}&\u{79.65} \\
		$\pkp=0.8$&77.97&93.31&74.82&\u{92.34}&72.31&90.75&56.98&79.48 \\        $\pkp=0.7$&76.90&\u{93.33}&\u{74.97}&92.32&\u{72.37}&\u{90.86}&56.97&79.48 \\
		$\pkp=0.6$&\u{77.01}&\b{93.45}&74.85&92.30&\b{72.41}&90.81&56.97*&79.35*  \\
		$\pkp=0.5$&\b{77.02}&\b{93.45}&\b{75.02}&\b{92.38}&72.22&90.69&56.59&79.11  \\  \midrule
	\end{tabular}
	\caption{\b{Stochastic Hide-and-Seek~\cite{singh2017hide} occlusions (joint training).}
		ImageNet top-1 and top-5 accuracies (\%) are averaged over 3 runs except where * (denotes 2 runs); stddev mean $= 0.14$ with range $[0.02, 0.31]$.
		ResNet50 notably improves by $0.62\%$ when jointly trained with H\&S occlusions. 
	}\label{t:hide_seek}
\end{table*}

\begin{table}[t]
	\centering
	\begin{tabular}{lccccccc}
		\toprule
		\multicolumn{1}{c}{} & \multicolumn{2}{c}{$S=$56} & \multicolumn{2}{c}{$S=$84} & \multicolumn{2}{c}{$S=$112} \\
		$N$&top-1&top-5&top-1&top-5&top-1&top-5\\ \midrule
		1&76.72&93.33&76.58&93.09&76.86&93.38\\
		2&76.55&93.12&76.94&93.32&\u{76.96}&\u{93.36}\\
		4&\u{76.82}&\u{93.47}&\u{77.07}&\u{93.47}&\b{77.19}&\b{93.46}\\
		6&\b{77.17}&\b{93.54}&\b{77.25}&\b{93.48}&76.80&93.39\\
		8&76.80&93.36&76.94&93.33&76.39&93.18\\
		\bottomrule
	\end{tabular}
	\caption{\b{Cutout for ResNet50 (joint training).}
		Results reported on one run.
		$S=$ side length of square patch; $N=$ \# of patches to cut out.
		For comparison, the joint baselines are $76.40\%$ (top-1) and $93.03\%$ (top-5) and the best joint Hide-and-Seek ($\pkp^*=0.5$) results are $77.02\%$ (top-1) and $93.45\%$ (top-5) from~\Cref{t:hide_seek}.}\label{t:cutout_grid_resnet_one_run}
\end{table}

\paragraph{Results.}~\Cref{t:hide_seek} reports ImageNet top-1 and top-5 accuracy for various networks when trained jointly with Hide-and-Seek occlusions, while~\Cref{t:cutout_grid_resnet_one_run} and~\Cref{t:cutout_grid_alexnet} reports results for ResNet50 and AlexNet respectively when trained jointly with Cutout occlusions.

\Cref{t:hide_seek} shows that ResNet50 improves significantly ($+0.62\%$ in top-1 and $+0.35\%$ in top-5 for the optimal $p_{\t{kp}}^* = 0.5$) when jointly trained with H\&S occlusions.
Furthermore, ResNet50 consistently beats the baseline (10 of 10 results improve) regardless of the $p_\text{keep\_patch}$ hyperparameter.
However, for all other networks, the best improvements are negligible: $0.07\%, 0.04\%, 0.02\%$ in top-1 and $0.07\%, -0.05\%, -0.07\%$  in top-5 for VGG16-BN, VGG16, and AlexNet respectively.
Consistent with the results reported in~\cite{hoffer2019augment}, the difference between the joint and non-joint baselines in~\Cref{t:hide_seek} appears roughly correlated with network performance, with ResNet50 having no difference and while the others demonstrate significant improvement with joint training: top-1 baselines improve by $0.00\%,  0.84\%, 0.62\%, 0.95\%$ for VGG16-BN, VGG16, and AlexNet respectively.

Thus, we focus our attention on ResNet50 for Cutout experiments.
~\Cref{t:cutout_grid_resnet_one_run} shows that Cutout with joint training on ResNet50 nearly always improves on the baseline (23 of 25 results improve), regardless of the size and number of patches occluded ($S$ and $N$).
The best result improves $0.85\%$ for top-1 and $0.38\%$ for top-5 over baselines, with the top-1 improvement being substantially higher with the best Cutout hyper parameters ($77.25\% \text{ with } N=6, S=84$) than that with the best Hide-and-Seek ones ($77.02\% \text{ with } \pkp=0.5$).

In contrast, \Cref{t:cutout_grid_alexnet} shows that Cutout with joint training on AlexNet rarely improves on the joint baseline (only 1 of 25 results improves; we include this table for comparison with saliency-based occlusions in~\Cref{s:saliency}).

Taken together, these results suggest that, for complex datasets like ImageNet, a suitably powerful architecture like ResNet50 is likely necessary to benefit from occlusion augmentation.

\paragraph{Occlusions as a stethoscope for model capacity.} The results for both kinds of stochastic occlusions (\Cref{t:hide_seek} and~\Cref{t:cutout_grid_resnet_one_run}) peak in performance with the best hyper-parameters and then roughly monotonically decrease from that point.
Thus, training with occlusions is beneficial from a model understanding perspective, as it provides a way to identify and quantify an architecture's upper bound for handling occlusions at evaluation time.
For Hide-and-Seek (\Cref{t:hide_seek}), we see that the optimal $\pkp^* \in [0.5, 0.6]$ for ResNet50 and VGG16-BN, $\pkp^* \in [0.6, 0.7]$ for VGG16, and $\pkp^* = 0.9$ for AlexNet.
This suggests that AlexNet can only handle a small amount of occlusion (images occluded up to 10\% on average), while VGG16-BN and ResNet50 are capable of handling images that have been occluded up to 50\% on average, when trained properly with occlusions (ResNet50 and VGG16-BN may be able to handle more than 50\%, but this was not tested).


\paragraph{Visualizations.}~\Cref{f:examples} compares a ResNet50 non-joint baseline against a ResNet trained jointly with Hide-and-Seek $\pkp=0.6$ best by using~\cite{rebuffi2019}'s saliency method on layer3.0.conv1.
Here, we visualize saliency maps for a few examples in which the occlusion-augmented network was correct and the baseline was wrong.
Qualitatively, we observe difference in the models' predictions in their visualizations:
In the suit image, the augmented network focuses on the tie while the baseline is attracted to the man's gaze and elbow.
Same with the ball player, we see the baseline's mistake in focusing on the bottom edge of the image.
In line with previous work~\cite{singh2017hide, wei2017object, wang2017fast}, we also observe that visualizations of the augmented network tend to cover the object surface more than those of the baseline model.

\begin{figure*}[t]
	\centering
	\includegraphics[width=0.32\linewidth]{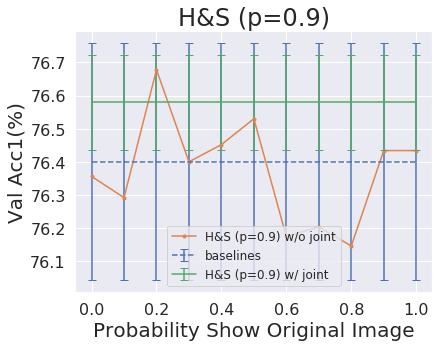}
	\includegraphics[width=0.32\linewidth]{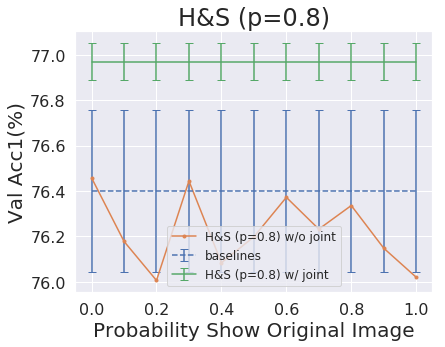}
	\includegraphics[width=0.32\linewidth]{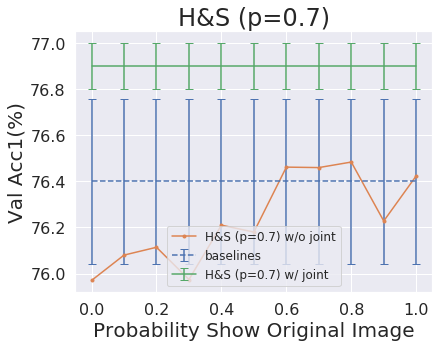}
	\includegraphics[width=0.32\linewidth]{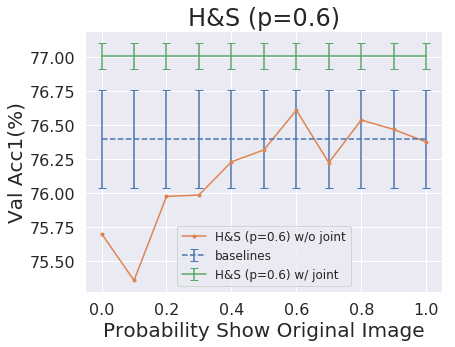}
	\includegraphics[width=0.32\linewidth]{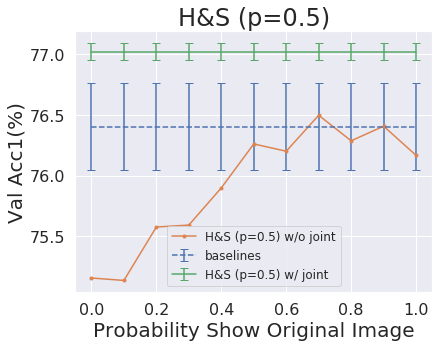}
	\includegraphics[width=0.32\linewidth]{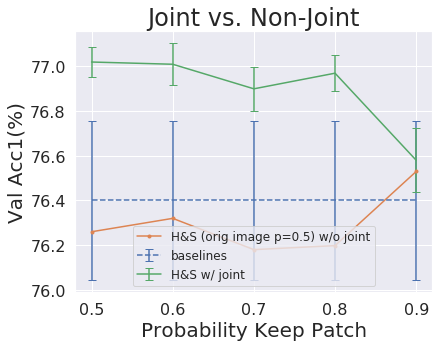}
	\caption{\b{Joint vs. Non-Joint Training on ResNet50.}
		We show that joint training (i.e., same image occluded and unoccluded in a mini-batch;  red lines) is necessary to improve over baselines (dotted lines) compared to leaving an image unoccluded randomly ($\pki$).
		All plots except the bottom right one show ResNet50 baselines and Hide-and-Seek (H\&S) joint and non-joint training for a given $\pkp$ as $\pki$ varies.
		The bottom right plot compares H\&S joint and non-joint training for $\pki=0.5$ as $\pkp$ varies.}\label{f:joint_vs_non_joint}
\end{figure*}

\subsection{Joint vs.~non-joint training}
We then thoroughly tested the necessity of joint training to make occlusion augmentation effective.
We trained networks with Hide-and-Seek occlusions \emph{without} joint training by introducing another hyper-parameter $\pki$ that determines whether an image is left completely unoccluded (see~\Cref{s:method} for more details).
We train these networks with $\pki = \{0.0, 0.1, \ldots, 1.0\}$ and $p_\text{keep\_patch} = \{0.5, 0.6, \ldots, 0.9\}$.
We then compare those networks with our baselines and our jointly trained networks from~\Cref{t:hide_seek}.
If joint training is not strictly necessary, we would expect our non-jointly trained networks to beat the baselines.

\Cref{f:joint_vs_non_joint} shows that this is not the case. Overwhelming, the non-jointly trained networks (green lines) perform worse than our baselines (dotted lines).
While we might expect that when $p_\text{keep\_patch}= 0$, that is, when images are always occluded and thus the training domain might be too different from the test domain, it is surprising that even when showing full images half of the time ($p_\text{keep\_patch} = 0.5$), we do not see an improvement.
This suggests that that seeing an image occluded and unoccluded \emph{in the same batch} is necessary for occlusion augmentation to work well.
Our finding are consistent with~\cite{ghiasi2018dropblock}'s observation that Cutout did not improve ImageNet classification performance.

We also briefly explored finetuning models on full images after they have been trained on exclusively occluded images but did not see an improvement over baselines.

\begin{table}[t]
	\centering
	\begin{tabular}{lccccccc}
		\toprule
		\multicolumn{1}{c}{} & \multicolumn{2}{c}{$S=$56} & \multicolumn{2}{c}{$S=$84} & \multicolumn{2}{c}{$S=$112} \\
		$N$&top-1&top-5&top-1&top-5&top-1&top-5\\ \midrule
		1&\u{57.32}&\u{79.58}&\b{57.24}&\b{79.65}&\b{57.32}&\b{79.66}\\
		2&\b{57.49}&\b{79.67}&\u{57.22}&\u{79.61}&\u{57.03}&\u{79.32}\\
		4&57.32*&79.57*&56.65&79.29&56.16&78.85\\
		6&56.94&79.37&56.39&78.84&55.37&78.16\\
		8&56.49&79.14&55.52&78.30&54.95&77.91\\
		\bottomrule
	\end{tabular}
	\caption{\b{Cutout for AlexNet (joint training).}
		Averaged over 2 runs except where * (denotes 1 run); standard deviation mean $= 0.92$ with range $[0.00, 0.32]$.
	}\label{t:cutout_grid_alexnet}
\end{table}

\begin{table}[t]
	\centering
	\setlength\tabcolsep{3pt}
	\begin{tabular}{lcccccc}
		\toprule
		\multicolumn{1}{c}{} & \multicolumn{2}{c}{VGG16-BN} & \multicolumn{2}{c}{VGG16} & \multicolumn{2}{c}{AlexNet} \\
		&top-1&top-5&top-1&top-5&top-1&top-5        \\ \midrule
		w/o joint&74.11&91.81&71.75&90.45&56.39&79.19 \\
		w/ joint&74.95&92.31&72.37&\b{90.91}&57.34&\u{79.72} \\ \midrule
		conv3&75.01&\b{92.41}&\b{72.48}&90.86&\b{57.42}&79.71 \\
		conv4&74.96&\u{92.40}&72.27&90.87&\u{57.38}&\b{79.74}\\
		conv5&\b{75.06}&92.39&72.38&\u{90.90}&57.33&79.70\\\midrule
		Best from Tbl~\ref{t:hide_seek}&\u{75.02}&92.38&\u{72.41}&90.86&57.36&79.65 \\
		\bottomrule
	\end{tabular}\label{t:saliency}
	\caption{\b{Saliency-based~\cite{rebuffi2019} occlusions for VGG16-BN, VGG16, and AlexNet (joint training).}
		Averaged over 3 runs; stddev mean $= 0.07$ with range  $[0.03, 0.16]$.
		Hyper-parameters $N=1$ occlusion, $S=56$ side length, $\tau=16$ jitter are used.}\label{t:saliency_most}
\end{table}

\begin{table}[t]
	\centering
	\begin{tabular}{lcc}
		\toprule
		\multicolumn{1}{c}{ResNet50} & top-1& top-5 \\ \midrule
		w/o joint&76.40\d&93.10\d \\
		w/ joint&76.40\d&93.03\d \\ \midrule
		maxpool&\underline{76.57}&\underline{93.19}\\
		layer1.0.conv1&76.21&93.06\\
		layer2.0.conv1&76.53&93.00\\
		layer3.0.conv1&76.36&93.12\\ \midrule
		Best from Tbl~\ref{t:hide_seek}&\b{77.02}\d&\b{93.45}\d \\
		\bottomrule
	\end{tabular}
	\caption{\b{Saliency-based occlusions for ResNet50 (joint training).}
		Results reported for 1 run (\d denotes averaged over 3 runs).
		These hyper-parameters were used: $N=1, S=56, \tau=16$.
	}\label{t:saliency_resnet}
\end{table}

\subsection{Saliency-based occlusions}\label{s:saliency}

\paragraph{Experiment set-up.} For AlexNet, VGG16, and VGG16-BN, we train networks with occlusions based on~\cite{rebuffi2019}'s saliency maps at the following layers (post-ReLU but pre-pooling): conv3, conv4, and conv5\footnote{For VGG16(-BN), convX refers to the last convolutional layer in the X-th block.}.
For ResNet50, we train networks on saliency maps on the max pool before the first block and on the very first convolutional layers in the first, second, and third blocks before batch normalization.
Given a saliency heatmap, we extract a $56 \times 56$ Cutout-like patch that covers the most salient part of the image.
We then jitter the patch uniformly by $\tau \in [-16, 16]$ pixels.

\paragraph{Results.}~\Cref{t:saliency_most} and~\Cref{t:saliency_resnet} show that the best results from training jointly with saliency-based occlusions for all networks except ResNet50 are consistently better (albeit by a small margin) than the best results from training jointly with stochastic Hide-and-Seek occlusions.
Most notably, a much smaller amount of saliency-based occlusion is needed to yield the comparable improvements to Hide-and-Seek occlusions (i.e., for VGG16-BN, occluding $6\%$ of an image using saliency is comparable to occluding $50\%$ on average using Hide-and-Seek).
This is likely due to the fact that the saliency-based occlusions should be covering the most ``important'' parts of an image.
Our saliency-based $N=1$ occlusion of side length $S=56$ is roughly comparable to Hide-and-Seek with a $4 \times 4$ grid ($G=4, S=56$) and $\pkp=15/16=0.94$, that is, on average only one $56 \times 56$ patch is occluded.
It is also is directly comparable with Cutout with the same hyper-parameters ($N=1, S=56$; see~\Cref{t:cutout_grid_resnet_one_run} and~\Cref{t:cutout_grid_alexnet} for Cutout on ResNet50 and AlexNet respectively).

The slim differences between results from different layers suggests that occlusions based on~\cite{rebuffi2019}'s saliency method are reasonably robust to layer choice.
Saliency-based occlusion also yields a lower mean standard deviation of $0.07$ compared to $0.14$ for Hide-and-Seek occlusions, due to the significantly less stochastic nature of saliency-based occlusion augmentation.

One limitation of our current approach is that we can extract one maximal patch, thereby limiting to a certain degree the size of our occlusions, which would need to be larger in order to match the effects of the best parameterizations of the stochastic methods.
This limitation is likely the reason that results from saliency-based $N=1$ occlusions on ResNet50 do not beat the best stochastic occlusion results, since a larger amount of occlusion is needed for Hide-and-Seek ($\pkp=0.5$) and Cutout ($N=6$ for $S = 56$).

\begin{table*}[t]
	\centering
	\begin{tabular}{lcccc}
		\toprule
		&\multicolumn{2}{c}{top-1 (\%)}&\multicolumn{2}{c}{top-5 (\%)} \\
		\multicolumn{1}{c}{ResNet50} &non-joint&joint&non-joint&joint \\ \midrule
		baseline from~\Cref{t:hide_seek}&\u{76.40}&76.40&\u{93.10}&93.03\\ \midrule
		Dropout~\cite{srivastava2014dropout} ($p_{\t{keep\_prob}}=0.9$)&76.34&76.41&93.02&93.10\\
		Spatial Dropout~\cite{tompson2015efficient} ($p_{\t{keep\_prob}}=0.9$)&75.95&76.31&92.77&93.04\\
		DropBlock~\cite{ghiasi2018dropblock} ($p_{\t{keep\_prob}}=0.95$ \& $p_{\t{keep\_prob}}=0.90$\d)&75.88&76.33&92.77&92.98\\ \midrule
		Label smoothing~\cite{szegedy2016rethinking} (0.1)&\b{76.64}&76.26&\b{93.25}&93.11\\ \midrule
		Best H\&S ($\pkp^*=0.5$) from~\Cref{t:hide_seek} (ours)&--&\u{77.02}&--&\u{93.45}\\
		Best CO ($N^*=6; S^*=84$) from~\Cref{t:cutout_grid_resnet_one_run} (ours)&--&\b{77.25}&--&\b{93.48}\\
		\bottomrule
	\end{tabular}
	\caption{\b{Comparison with other regularization methods.}
		For Dropout variants, the best results from a search over several $p_{\t{keep\_prob}}$ values is reported.
		\d For DropBlock, $p_{\t{keep\_prob}}=0.95$ was the best for non-joint and $p=0.90$ was best for joint.
	}\label{t:regularization_comparison}
\end{table*}

\subsection{Comparison with other regularization methods}
\paragraph{Experimental set-up.} We compare our method with variants of Dropout~\cite{ghiasi2018dropblock} and primarily follow~\cite{ghiasi2018dropblock}'s protocol (see~\Cref{s:related} for more details).
For Dropout~\cite{srivastava2014dropout}, Spatial Dropout~\cite{tompson2015efficient}, and DropBlock~\cite{ghiasi2018dropblock}, we follow~\cite{ghiasi2018dropblock}'s procedure and add dropout modules after every convolutional layer in the third and fourth block of ResNet50.
For DropBlock, we also add its module to the skip connections in those blocks.
For Dropout and Spatial Dropout, we train ResNet50 networks without joint training using $p_{\t{keep\_prob}} \in \{0.5, 0.6, 0.7, 0.8, 0.9\}$, while for DropBlock, we use $p_{\t{keep\_prob}} \in \{0.75, 0.80, 0.85, 0.90, 0.95\}$ ($p_{\t{keep\_prob}}$ is analogous to $\pkp$).
We also compare against label smoothing~\cite{szegedy2016rethinking} with fixed $p=0.1$.

We deviate from~\cite{ghiasi2018dropblock} in that we train for 100 epochs using a 30--60--90 epoch lr decay schedule (vs.~their 300 epochs using a 100--200--265 schedule) to compare fairly with our method.
We do not use a schedule to ease in the amount of dropout for DropBlock, as~\cite{ghiasi2018dropblock} reported that DropBlock without scheduling still yielded significant boosts over their ResNet50 baseline.
We expected that these two changes would decrease the improvements observed in~\cite{ghiasi2018dropblock} but that those improvements would still persist.

\paragraph{Results.} \Cref{t:regularization_comparison} shows that all the variants of Dropout methods under-performed our ResNet50 non-joint training baseline, suggesting that they are sensitive to and require the custom longer training schedule used in~\cite{ghiasi2018dropblock} in order to be effective (see~\cite{ghiasi2018dropblock} for results with the longer training schedule).
Label smoothing also under-performed our occlusion augmentation training.

\begin{table*}[t]
	\centering
	\begin{tabular}{lcccc|cccccc}
		\toprule
		\multicolumn{1}{c}{}&\multicolumn{2}{c}{JT bsl (Tbl~\ref{t:hide_seek})} & \multicolumn{2}{c}{JT CO (Tbl~\ref{t:cutout_grid_resnet_one_run})} &\multicolumn{2}{c}{BA bs}&\multicolumn{2}{c}{BA CO ($p_{\t{ki}}=0.0$)}&\multicolumn{2}{c}{BA CO ($p_{\t{ki}}=0.5$)}\\
		$M$&top-1&top-5&top-1&top-5&top-1&top-5&top-1&top-5&top-1&top-5\\\midrule
		2&76.40\d&93.03\d&77.17*&93.54*&77.27*&93.47*&\b{77.50}*&\b{93.63}*&\b{77.50}&93.61\\
		4&--&--&--&--&77.71*&\b{93.79}*&77.75&93.68&\b{77.82}&93.74\\
		\bottomrule
	\end{tabular}
	\caption{\b{Batch augment (BA) vs.~joint training (JT) for Cutout (CO) on ResNet50.}
		Averaged over 2 runs except where * (1 run) and \d (3 runs); stddev mean $= 0.09$ and range $[0.02, 0.18]$ for 2-run results.
		Best results \emph{per row} are in bold.
		CO hyper-parameters were $S=56$ and $N=6$.
		$M=$ \# copies of an image in a mini-batch.
		$p_{\t{ki}} = p_{\t{keep\_image}}$.
	}\label{t:augment_cutout}
\end{table*}

\begin{table}[t]
	\centering
	\setlength\tabcolsep{3pt}
	\begin{tabular}{lcccccc}
		\toprule
		\multicolumn{1}{c}{} &\multicolumn{2}{c}{Batch Augment}&\multicolumn{2}{c}{Dataset Augment}&\multicolumn{2}{c}{Joint Training}\\
		$M$&top-1&top-5&top-1&top-5&top-1&top-5\\\midrule
		2&\b{77.27}&\b{93.47}&76.51&93.12&76.39&93.14\\
		4&\b{77.71}&\b{93.79}&76.01&92.53&76.30&93.07\\
		\bottomrule
	\end{tabular}
	\caption{\b{Baseline comparisons with different kinds of augmentation.} Results are from 1 run. $M=$ number of copies of an image in a mini-batch for BA and JT and in one epoch of training for DA.}\label{t:augment_vs_joint}
\end{table}

\subsection{Comparison with Batch Augmentation~\cite{hoffer2019augment}}\label{s:augment}

Lastly, we compare our joint training paradigm with batch augmentation~\cite{hoffer2019augment}.
The key difference between batch augmentation and joint training is that, for joint training, all standard pre-processing occurs \emph{before} image duplication; in contrast, for batch augmentation, pre-processing occurs \emph{after} duplication.
Thus, transformation from pre-processing are \emph{identical} in joint training but independent (i.e., different) in batch augmentation. In all our previous results, we used joint training ($M=2$ copies).

\paragraph{Batch augment $>$ joint training.}~\Cref{t:augment_cutout} shows results when we use batch augmentation to include stochastic Cutout occlusions during training, with fixed CO hyper-parameters $N=6, S=56$.
The results for $M=2$ in~\Cref{t:augment_cutout} improve upon and are comparable to our joint training Cutout results for $N=6, S=56$ in~\Cref{t:cutout_grid_resnet_one_run}: $77.50\%$ (top-1) and $93.63\%$ (top-5) for batch augmented Cutout ($\pki=0.5$\footnote{$\pki$ denotes the probability that an image copy is left unoccluded.}) vs. $77.17\%$ (top-1) and $93.54\%$ (top-5) for joint training.
However, batch augmentation also significantly improves its respective baseline; thus, relative improvement of batch-augmented Cutout are smaller when compared to that of jointly trained Cutout:
For $M=2$, is quite slim for top-1 (and non-existent for top-5) when using $M=4$ copies.

\paragraph{No full images needed.} Most notably, Cutout with $\pki=0.0$ achieves similar performance to that with $\pki=0.5$.
This suggests that one can train a network with images that are \emph{always occluded} (i.e., without ever seen a full, natural image) and achieve superior inference-time performance on full images than standard training methods.

\paragraph{Baseline comparisons.}~\Cref{t:augment_vs_joint} shows results when training baseline ResNet50 models with batch augmentation, dataset augmentation, and joint training.
Dataset augmentation iterates through the training set $M$ times (i.e., $M$ copies are in \emph{distinct} mini-batches), while batch augmentation copies an image $M$ in the same mini-batch.
These results verify~\cite{hoffer2019augment} by showing the necessity of having image in the \emph{same} mini-batch.

\section{Conclusion}\label{s:conclusion}

We show an effective paradigm for using occlusion augmentation to improve ImageNet classification performance.
The primary insight from our work is using some variant of batch augmentation~\cite{hoffer2019augment} is necessary to gain this improvement.
This suggests that further research on what is being learned during joint training and more broadly batch augmentation~\cite{hoffer2019augment} is warranted.
We also demonstrate training-time occlusions can be a way to understand model's upper bound for robustness to occlusions generally.
There is likely room to improve our work here, particularly in exploring further the potential of batch augmentation~\cite{hoffer2019augment}, in developing better saliency-based approaches to occlusion augmentation, and in elucidating further the interaction between and impact of dataset and model complexity for effective occlusion augmentation.
Further research could also be done on other kinds of occlusions, such as blur or random noise or even ignoring regions~\cite{liu2018image}.
In conclusion, in contrast to other regularization techniques that require architectural changes, we present a simple paradigm for making occlusions effective on ImageNet for sufficiently capable models (e.g., ResNet50) that can be easily added into existing training paradigms.
\blfootnote{\noindent\textbf{Acknowledgements.}
	We are grateful for support from Open Philanthropy Project (R.F.). We also thank Sylvestre-Alvise Rebuffi for helpful discussions and sharing his code as well as Chris Olah and the OpenAI Clarity team for organizing insightful discussions on interpretability research.}

{\small\bibliographystyle{ieee}\bibliography{refs}}
\end{document}